# *Ro*botic Pollination of Apples in Commercial Orchards

Ranjan Sapkota[1,*], Dawood Ahmed[1], Salik Ram Khanal[1], Uddhav Bhattarai[1], Changki Mo[2], Matthew D. Whiting[3] and Manoj Karkee[1,*]

[1]Department of Biological Systems Engineering, Center for Precision and Automated Agricultural Systems, Washington State University, Prosser, WA, 99350, USA
[1]School of Mechanical and Materials Engineering, Washington State University Tri-Cities, Richland 99354, WA, USA
[3]Department of Horticulture, Washington State University, Prosser, WA, 99350, USA
*Contact: manoj.karkee@wsu.edu*

*Abstract*— This research presents a novel, robotic pollination system designed for targeted pollination of apple flowers in modern fruiting wall orchards. Developed in response to the challenges of global colony collapse disorder, climate change, and the need for sustainable alternatives to traditional pollinators, the system utilizes a commercial manipulator, a vision system, and a spray nozzle for pollen application. Initial tests in April 2022 pollinated 56% of the target flower clusters with at least one fruit with a cycle time of 6.5 s. Significant improvements were made in 2023, with the system accurately detecting 91% of available flowers and pollinating 84% of target flowers with a reduced cycle time of 4.8 s. This system showed potential for precision artificial pollination that can also minimize the need for labor-intensive field operations such as flower and fruitlet thinning.

## I. INTRODUCTION

The rapid decline of bee populations worldwide poses a critical threat to global food production systems [1]. Bees, as natural pollinators, are integral to agriculture, contributing to an estimated $235 billion in annual global crop production [2]. However, challenges such as global warming, agricultural practices, and Colony Collapse Disorder (CCD) are severely diminishing bee populations, placing a significant burden on agricultural yield and food security [3].

Worldwide, 33% of the food we consume is a direct result of pollination by bees. The declining bee populations thus represent a looming crisis for global food production and nutrition [4]. Current alternatives, such as manual hand pollination, are labor-intensive and economically unfeasible on a large scale, further emphasizing the need for an effective, scalable solution.

In response to the urgent issue of declining natural pollinators, this research introduces a novel, robotic system designed for the autonomous pollination of targeted flowers (the ones with the potential to develop good quality fruit in the desired canopy locations) in commercial apple orchards. Efficient pollination of target flowers promotes a one-fruit-per-flower-cluster outcome, which is regarded as an optimal approach for crop load management in modern apple orchards. This strategic focus enhances both the quality and efficacy of apple production, as it naturally thins fruit clusters and reduces the need for manual labor later in the season. Our system integrates advanced computer vision, GPS navigation, and robotic manipulation technologies to precisely identify and pollinate the targeted flowers, thereby reducing dependency on both natural pollinators and human labor. This extended abstract summarizes the findings from our field tests demonstrating the system's capability to achieve a high rate of pollination under real-world conditions, making it a viable and sustainable alternative for commercial adoption.

## II. METHODOLOGY AND SYSTEM DESCRIPTION

Three primary subsystems are comprised in the autonomous pollination robot developed in this study: i) a Machine Vision System, ii) a Manipulation System, and 3) an End Effector System. Each of these subsystems was designed to play a crucial role in enabling the robot to identify and pollinate targeted king flowers effectively and efficiently. Below, a brief description of each subsystem is provided.

### A. Machine Vision System

To enable the accurate identification of target flowers in natural orchard conditions, an advanced machine vision subsystem was employed in this system. An Intel RealSense camera was used to acquire images, which were then processed with a state-of-the-art deep-learning model to detect and locate the target flowers [5].

### B. Manipulation System

Once the target flowers were identified by the vision system, a commercially available manipulator was used to position the pollination end-effector to the desired location before spraying the pollens. The manipulator used had 6 DOF with a mounting mechanism in the last link, which provided the desired dexterity and precision to reach the target locations in the fruiting wall canopies and position the end-effector in the desired orientations.

### C. End Effector System

The terminal component of the robot was the end effector system that was needed to complete the pollination process. A spray nozzle was retrofitted and mounted onto the robotic arm for this purpose. This sprayer was supplied with a proprietary mixture of pollen, water, and a specialized pollen suspension medium. When activated, the solution was discharged onto the target flowers by the sprayer. In this system, the nozzle was positioned at a specific distance (25 cm) from the center of the target flowers (determined by the vision system), and the solution was sprayed for a specific duration.

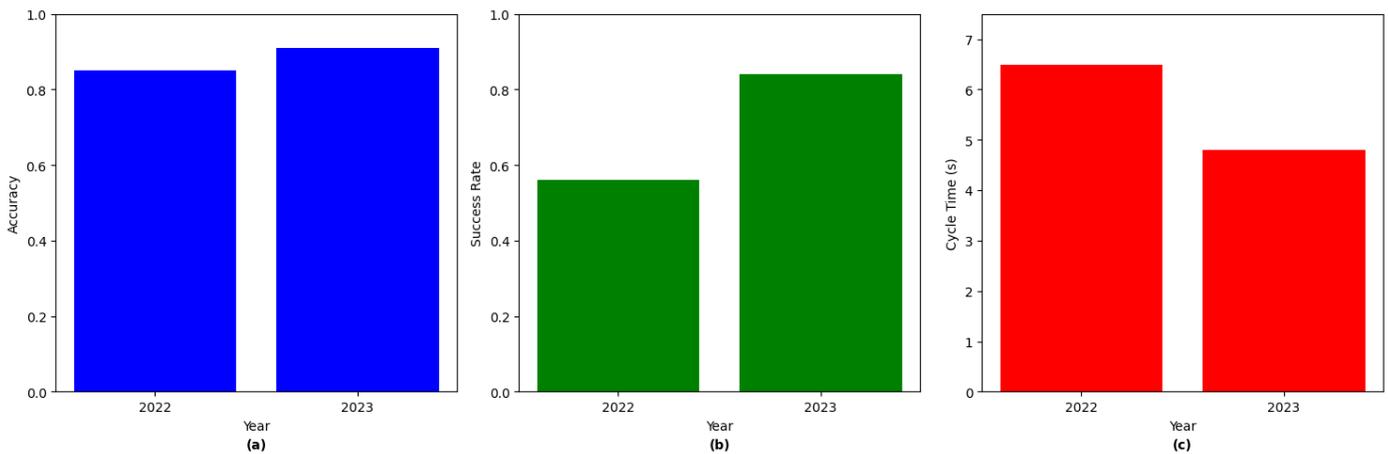

Figure 1: Performance of the pollination robot in 2022 and 2023 studies; (a) Flower Detection Accuracy; (b) Pollination Success Rate (%); and (c) Cycle Time. Notable improvements in flower detection and pollination success, as well as a reduction in cycle time, were observed in 2023 compared to the same achieved in 2022.

## III. RESULTS

A field study was conducted in a commercial Envy apple orchard to assess the performance of the robotic pollination system. In the study, the robot was driven between the tree rows with manual control, whereas the target flowers were identified automatically using the integrated machine vision system. The manipulator arm was then maneuvered to position the spray nozzle appropriately, and the pollen solution was subsequently sprayed onto the target flowers.

The vision system achieved 91% accuracy in detecting target flowers in complex orchard environments with the presence of a large number of non-target flowers, which showed that the machine vision system is reliable under diverse and dynamic environmental conditions encountered in commercial orchards. A primary metric evaluated during these field studies was the pollination success rate. The system achieved a success rate of 84% in pollinating the target flowers detected by the vision system. The average cycle time, defined as the time taken for a king flower to be detected and localized by the robot, maneuvered to, and pollinated, was recorded at 4.8 s. Figure 1 shows the performance comparison of the autonomous pollination robot between the years 2022 and 2023. The subplots show (a) Flower Detection Accuracy, (b) Successful Pollination Rate, and (c) Cycle Time for each year. Notable improvements in flower detection and pollination success rates, as well as a reduction in cycle time, are observed from 2022 to 2023.

In the orchard plots targeted by the robotic pollinator, the yield data showed promising results. The results suggested that the precision pollination of target flowers in apple orchards could be possible with a robotic system, which has the potential to streamline orchard management by reducing the biological and environmental variability and uncertainty in pollination while minimizing the need for later-stage flower and fruitlet thinning practices.

## IV. CONCLUSIONS AND FUTURE WORK

This study successfully demonstrated the potential of a novel, robotic pollination system as a viable alternative for precision pollination of apples in commercial orchards. With significantly improved detection accuracy, pollination success rate, and operational efficiency observed in our field trials in 2023 compared to the same in 2022 (Figure 1), this system provided a foundation for further advancing and potentially commercializing robotic pollination systems in apple orchards.